\documentclass[runningheads]{llncs}

\usepackage{eccv}

\usepackage{eccvabbrv}

\usepackage{makecell}

\usepackage{graphicx}
\usepackage{booktabs}

\usepackage[accsupp]{axessibility}  

\usepackage{hyperref}

\usepackage{orcidlink}

\usepackage{amsmath}
\usepackage{amssymb}
\usepackage{xcolor}
\usepackage{multirow}
\usepackage{booktabs}
\usepackage{algorithm}
\usepackage{algorithmic}

\begin{document}

\title{4D Reconstruction with Events \& RGB}


\author{
Dulanga Weerakoon\inst{1} \and
Darshana Rathnayake\inst{1} \and
Van Minh Do\inst{1} \and
Hemanth Sabbella\inst{3} \and
Kaichen Zhou\inst{2} \and
Fangneng Zhan\inst{2} \and
Alok Prakash\inst{1} \and
Sanjay E. Sarma\inst{2} \and
Archan Misra\inst{3} \and
Paul Pu Liang\inst{2}
}

\authorrunning{D.~Weerakoon et al.}
\institute{Singapore-MIT Alliance for Research \& Technology Centre \and Massachusetts Institute of Technology \and Singapore Management University}

\maketitle

\begin{figure*}
    \centering
    \includegraphics[width=1\linewidth]{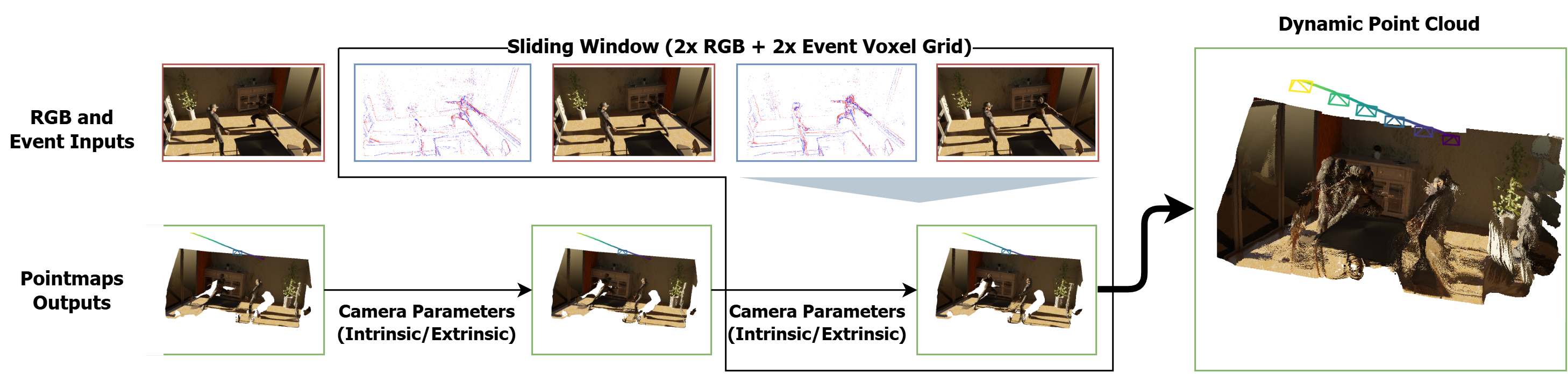}
    \caption{Overview of our event-augmented 4D reconstruction pipeline. Pairs of RGB frames and their corresponding event voxel grids are processed within a sliding window and passed through MonST3R to predict per-window pointmaps. Camera intrinsics and extrinsics estimated from overlapping windows align consecutive pointmaps into a single coordinate system, yielding a dynamic point cloud that captures scene geometry and motion over time.}
    \label{fig:teaser}
\end{figure*}

\begin{abstract}
\footnotetext[1]{Accepted at NeVi 2026 workshop at the 19th European Conference on Computer Vision (ECCV 2026).}
Reconstructing dynamic 3D scenes over time (4D reconstruction) from a single moving camera is central to robotics and AR/VR applications, yet state-of-the-art feed-forward methods rely solely on RGB frames and therefore inherit the failure modes of conventional cameras, including motion blur and temporal aliasing under fast or non-uniform camera and object motion. Event cameras report per-pixel brightness changes asynchronously at microsecond resolution, offering a complementary signal that is largely immune to these artifacts, but whether this modality can be incorporated into existing RGB-based 4D reconstruction pipelines without a full architectural redesign remains an open question. In this preliminary study, we investigate a minimal extension: we augment Monst3R, an existing RGB-only 4D reconstruction model, with an additional event-camera input branch, leaving the rest of the architecture unchanged. Across 3 benchmark datasets, this simple RGB-event fusion reduces Abs Rel by 3.4\% on average relative to the RGB-only Monst3R baseline, and a sensitivity analysis shows that the improvement grows with the magnitude of camera and object motion, with the largest gains observed under high object motion conditions where RGB-only reconstruction degrades most. These early results suggest that even a lightweight, non-invasive fusion of event data can meaningfully improve 4D reconstruction robustness, motivating a more thorough architectural investigation in future work. 

\end{abstract}

\section{Introduction}
\label{ref:intro}

Reconstructing dynamic three-dimensional scenes over time, referred to as \emph{4D reconstruction}, is a foundational problem in computer vision with applications in AR/VR content creation, robotics, autonomous driving, and digital twins. Recent advances in vision-language models (VLMs), such as CL4D~\cite{hewagamage2026cl4d}, have also begun exploring dynamic 4D point cloud representations as an alternative to conventional 2D video inputs, enabling richer spatio-temporal and geometric reasoning and further motivating accurate 4D reconstruction directly from videos. Recent feed-forward models, such as MonST3R~\cite{zhang2025monst3r}, recover per-frame depth and camera pose directly from RGB video without per-scene optimization. However, because these models rely solely on RGB frames, they inherit the failure modes of conventional cameras: motion blur and temporal aliasing under fast or non-uniform camera and object motion, precisely the conditions where accurate motion estimation matters most.

Event cameras offer a complementary sensing modality. By reporting per-pixel brightness changes asynchronously at microsecond resolution, they capture motion information largely unaffected by blur or frame-rate limitations. Prior work has explored event-based depth and pose estimation as standalone tasks, but whether an event stream can be incorporated into an existing RGB-based 4D reconstruction pipeline without a full architectural redesign remains less explored.

In this preliminary work, we present an early exploration of event-augmented 4D reconstruction through a minimal extension to the MonST3R pipeline as shown in Figure~\ref{fig:teaser}: we augment MonST3R with an additional event-voxel input branch, leaving the rest of the architecture unchanged, and fuse it with RGB features via concatenation before processing through the standard MonST3R pipeline. We hypothesize that RGB and event data are complementary---RGB provides spatial information for depth, while events provide motion cues for modeling dynamic scenes, particularly under fast object and camera motion---and that even this minimalist, concatenation-based fusion, without a more principled fusion mechanism, is sufficient to realize this complementarity.

To test this hypothesis, following prior work, we evaluate on benchmark datasets for both camera pose estimation and video depth estimation. We observe that fusing event voxels and RGB achieves higher reconstruction accuracy than using either modality alone, and that this benefit is particularly pronounced under high object motion.

While this work is a preliminary study on 4D reconstruction with RGB and events, we argue that more principled multi-modal fusion mechanisms are needed to achieve substantially higher reconstruction accuracy, given the fundamental differences between the two modalities: RGB is a constant-frame-rate, synchronous modality, while events are adaptive and asynchronous.

We make the following key contributions:

\begin{itemize}
    \item We extend MonST3R with a minimal event-voxel input branch, requiring no other architectural changes, and show that fusing RGB and event voxels reduces depth and pose estimation error relative to RGB-only MonST3R, an event-to-RGB translation baseline, and an event-only variant.
    \item We conduct a sensitivity study under varying camera and object motion, establishing that RGB-event fusion provides the largest gains under high object motion.
\end{itemize}

\section{Related Work}
\label{ref:related}

\noindent\textbf{Feedforward 4D reconstruction.}
Prior work on 4D reconstruction primarily used RGB frames as input. Early work such as DUSt3R~\cite{wang2024dust3r} proposed predicting a separate pointmap for each pair of RGB frames. MonST3R~\cite{zhang2025monst3r} extended the DUSt3R~\cite{wang2024dust3r} pointmap paradigm to dynamic scenes. Both MonST3R and DUSt3R predict pointmaps followed by a postprocessing step that aligns them into a single coordinate system. Works such as CUT3R~\cite{wang2025continuous} and Fast3R~\cite{yang2025fast3r} instead unify pointmap prediction and alignment into a single feed-forward pipeline, directly predicting aligned pointmaps in one coordinate system without a separate postprocessing step. More recently, VGGT~\cite{wang2025vggt} and PAGE4D~\cite{zhou2026page} scale up training data and model capacity with transformer-based architectures, achieving more accurate 3D/4D scene reconstruction.

Despite this large body of work on 4D reconstruction, all prior methods take a single monocular RGB video as input and are feed-forward, and none incorporate an event stream. Robustness to fast motion remains an open challenge across these methods, motivating the present work: fusing event and RGB data for 4D reconstruction.

\noindent\textbf{Event cameras and RGB event fusion.} Event cameras have been studied extensively for their high temporal resolution, negligible motion blur, and high dynamic range~\cite{gallego2020event}. RGB and event fusion has been explored successfully for related but narrower tasks, including video deblurring~\cite{kim2024frequency}, frame interpolation/extrapolation~\cite{tulyakov2021time, rathnayake2026neurolidar}, and event-based visual odometry and SLAM~\cite{vidal2018ultimate}, where events are typically used to recover information that has already been lost in blurred or dropped RGB frames. Its potential for 4D scene reconstruction, that is, jointly recovering time-varying geometry and appearance rather than a single deblurred or interpolated frame, remains comparatively underexplored.
 
\noindent\textbf{Event-augmented 4D reconstruction.} More recently, EAG3R~\cite{wu2026eag3r} presented the first event-augmented 4D reconstruction method, combining RGB and event data to target low-light conditions. Its design and evaluation are organized around illumination robustness: an SNR-aware fusion module favors RGB features in well-lit regions and event features in dark regions, and its central experiments vary illumination from day to night while holding the motion regime relatively constant. This leaves comparatively less direct evidence for a complementary motivation for event cameras: their ability to capture fast, non-rigid scene dynamics that a fixed-frame-rate RGB camera under-samples or blurs, independent of illumination. The present work targets precisely this gap, evaluating explicitly under varying camera and object motion rather than varying illumination.

\section{Methodology}
\label{sec:method}

This section describes our approach for 4D reconstruction using multi-modal RGB and event inputs. We first introduce the event representation, followed by the proposed network architecture and the dataset preparation pipeline.

An asynchronous event stream over the time interval $(t, t+\Delta)$ is defined as
\begin{equation}
\mathcal{E}_{(t, t+\Delta)} = \{e_i = (p_i, x_i, y_i, t_i)\}_{i=0}^{N},
\end{equation}
where $p_i \in \{-1,1\}$ denotes the event polarity corresponding to an increase or decrease in brightness, $(x_i, y_i) \in \mathbb{R}^{H \times W}$ represents the pixel coordinates, and $t_i$ denotes the event timestamp.

Our framework builds upon the MonST3R architecture and extends it to support multi-modal inputs consisting of RGB images and event data. The network takes two RGB images, $I_1, I_2 \in \mathbb{R}^{W \times H \times 3}$, together with two event voxel grids, $\mathcal{E}_1, \mathcal{E}_2 \in \mathbb{R}^{W \times H \times B}$, where $B=5$ denotes the number of temporal bins. Here, $H$ and $W$ represent the image height and width, respectively.

The event stream is converted into a voxel grid representation~\cite{zhu2019unsupervised} by accumulating events within each temporal bin as
\begin{equation}
\mathcal{E}_{V}(b,x,y)=
\sum_{e_i \in \mathcal{E}_{(t,t+t_1)}} p_i
\mathbf{1}
\left\{
b=\left\lfloor
B \cdot \frac{t_i-t}{t_1}
\right\rfloor,
x=x_i,
y=y_i
\right\},
\label{eq:voxel}
\end{equation}
where $\mathbf{1}\{\cdot\}$ is the indicator function.

\subsection{Model Architecture}

We extend MonST3R to jointly process RGB images and event voxel grids while preserving the original output representations. For each view, the 3-channel RGB image and the 5-channel event voxel grid are concatenated to form an 8-channel input tensor. Since the pretrained ViT encoder expects a 3-channel input, we introduce a lightweight adapter consisting of a single convolution layer that projects the 8-channel input to a 3-channel feature representation. The adapted inputs from the two views are then processed by the shared Siamese ViT encoder, followed by the original MonST3R decoder and prediction heads without further architectural modifications.

\subsection{Dataset Preparation}
\label{sec:dataset-prep}

The datasets used in this work provide only RGB image sequences. To obtain the corresponding event data, we use the V2E~\cite{hu2021v2e} simulator to generate asynchronous event streams between consecutive RGB frames. The simulator first interpolates intermediate RGB frames, effectively increasing the temporal resolution of the original video sequence. It then generates events by modeling the brightness changes between the interpolated frames. Finally, the generated event streams are accumulated into voxel grid representations according to Equation~\ref{eq:voxel}, which are used together with the corresponding RGB images as inputs to the proposed network.

\subsection{Training Details}

Similar to MonST3R, we train the proposed variants on a number of prior datasets, each augmented with synthetically generated event data as described in Section~\ref{sec:dataset-prep}: PointOdyssey~\cite{zheng2023point}, TartanAir~\cite{wang2020tartanair}, and Spring~\cite{mehl2023spring}. Remaining training configurations are identical to the original MonST3R implementation.

\section{Results}
\label{ref:results}

\subsubsection{Evaluation Datasets.} We evaluate on Sintel~\cite{Butler:ECCV:2012}, Bonn~\cite{palazzolo2019iros}, KITTI~\cite{geiger2013vision}, and PointOdyssey~\cite{zheng2023point}, and adopt the standard evaluation metrics used in the 4D reconstruction literature, following MonST3R~\cite{zhang2025monst3r}.

\subsubsection{Baseline.} We primarily use MonST3R as the base model with different input modalities. (a) RGB is the MonST3R model with the pretrained weights from the original paper implementation. (b) RGB (HyperE2VID) also uses the pretrained MonST3R weights, but RGB frames are generated from event voxels using HyperE2VID~\cite{ercan2024hypere2vid} instead of the sensor's original RGB frames. (c) Ev. Vox. retrains MonST3R using event voxels (bin size = 5) alone, without RGB input. (d) RGB + Ev. Vox. retrains MonST3R using concatenated RGB and event voxels (bin size = 5).

\subsubsection{Camera Pose Estimation.} Table~\ref{tab:pose-sintel} compares camera pose estimation error on Sintel across four input variants to MonST3R. RGB combined with event voxels achieves the lowest error on all three metrics (ATE mean, RPE trans mean, RPE rot mean). Reconstructing RGB from events via HyperE2VID before passing it to MonST3R degrades performance relative to using the original RGB frames directly, indicating that direct fusion of event voxels is more effective than a translation-based approach, which may inherently introduce errors during event-to-RGB conversion. Event voxels alone generally achieve a higher error than RGB + Event voxels, underscoring the importance of fusing RGB and event data for accurate depth reconstruction.

\subsubsection{Video Depth Estimation.} Table~\ref{tab:main-results} reports video depth estimation error on three datasets: Sintel, Bonn, and KITTI. As with camera pose estimation, RGB combined with event voxels achieves the lowest error on both metrics across all three datasets. A MonST3R variant trained with event voxels alone achieves the highest error, exceeding even the translation-based approach of converting events to RGB with an event-to-RGB model before passing them to MonST3R. This further reinforces that spatial information from RGB, coupled with motion-specific cues from event data, is necessary for accurate 4D reconstruction.

\begin{table}[t]
\centering
\caption{Camera pose estimation on Sintel with MonST3R as the base model. \textbf{Bold} indicates the best result per column.
$\downarrow$: lower is better.}
\label{tab:pose-sintel}
\begin{tabular}{lccc}
\toprule
Input & ATE mean $\downarrow$ & RPE trans mean $\downarrow$ & RPE rot mean $\downarrow$ \\
\midrule
RGB & 0.108 & 0.042 & 0.732 \\
RGB (HyperE2VID)    & 0.132 & 0.067 & 0.767 \\
\midrule
Ev. Vox.    & 0.154 & 0.076 & 0.532 \\
RGB + Ev. Vox.    & \textbf{0.106} & \textbf{0.038} & \textbf{0.476} \\
\bottomrule
\end{tabular}
\end{table}

\begin{table}[t]
\centering
\caption{Video depth estimation comparison across three datasets with MonST3R as the base model. \textbf{Bold} indicates
the best result per column. $\downarrow$: lower is better, $\uparrow$: higher is better.}
\label{tab:main-results}
\begin{tabular}{l cc|cc|cc}
\toprule
& \multicolumn{2}{c}{Sintel} & \multicolumn{2}{c}{Bonn} & \multicolumn{2}{c}{KITTI} \\
\cmidrule(lr){2-3} \cmidrule(lr){4-5} \cmidrule(lr){6-7}
Input & Abs Rel $\downarrow$ & $\delta{<}1.25 \uparrow$ & Abs Rel $\downarrow$ & $\delta{<}1.25 \uparrow$ & Abs Rel $\downarrow$ & $\delta{<}1.25 \uparrow$ \\
\midrule
RGB & 0.345 & 56.2 & 0.065 & 96.3 & 0.106 & 89.3 \\
RGB (HyperE2VID) & 0.464 & 53.3 & 0.085 & 93.3 & 0.143 & 79.9 \\
\midrule
Ev. Vox.    & 0.47 & 50.2 & 0.088 & 92.5 & 0.165 & 78.3 \\
RGB + Ev. Vox.    & \textbf{0.331} & \textbf{58.1} & \textbf{0.064} & \textbf{96.4} & \textbf{0.101} & \textbf{90.6} \\
\bottomrule
\end{tabular}%
\end{table}

\begin{figure}[t]
\centering
\begin{subfigure}[b]{0.48\linewidth}
    \centering
    \includegraphics[width=\linewidth, height=1.5in]{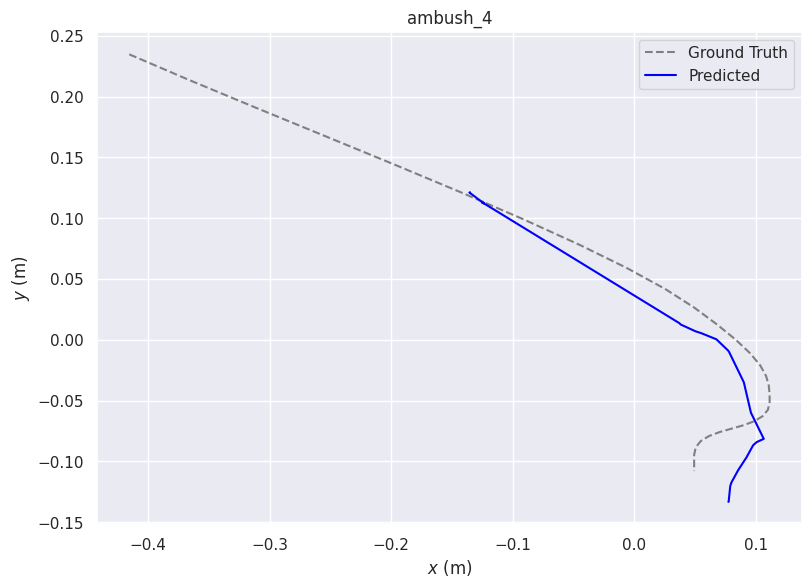}
    \caption{ambush\_4}
    \label{fig:traj-ambush6}
\end{subfigure}
\hfill
\begin{subfigure}[b]{0.48\linewidth}
    \centering
    \includegraphics[width=\linewidth, height=1.5in]{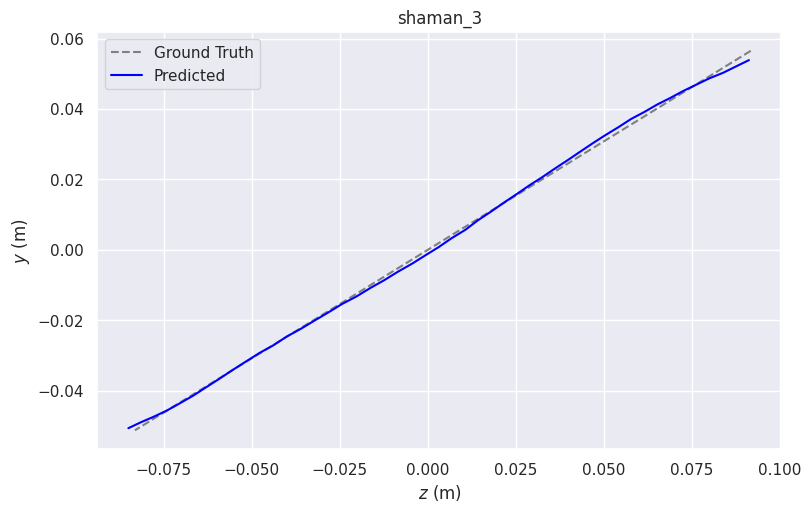}
    \caption{shaman\_3}
    \label{fig:traj-shaman3}
\end{subfigure}
\caption{Predicted vs.\ ground-truth camera trajectories on two Sintel sequences with RGB + Ev. Vox.}
\label{fig:trajectories}
\end{figure}

\begin{figure}[t]
\centering
\begin{subfigure}[b]{0.48\linewidth}
    \centering
    \includegraphics[width=\linewidth]{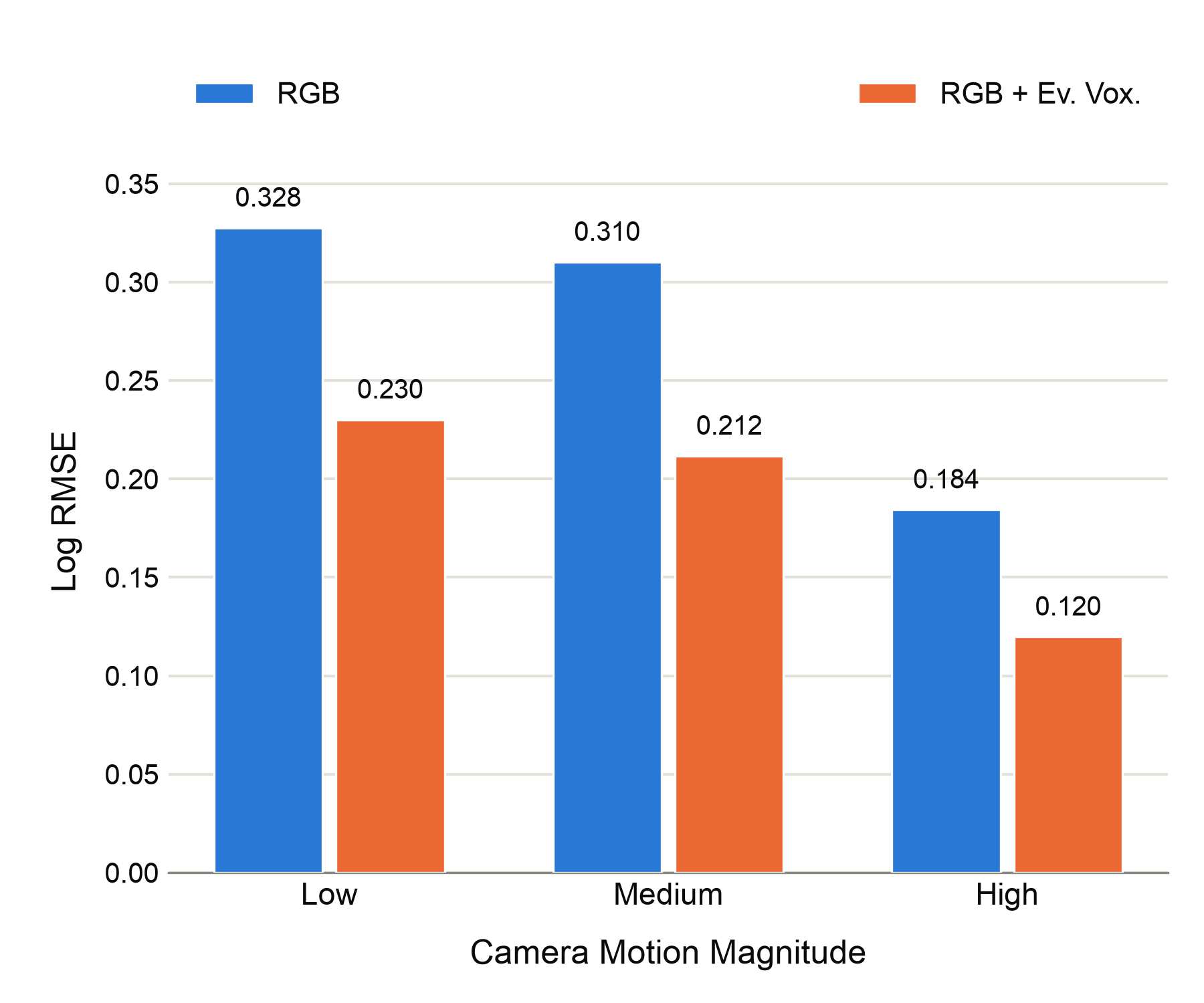}
    \caption{Camera motion}
    \label{fig:sens-cam-motion}
\end{subfigure}
\hfill
\begin{subfigure}[b]{0.48\linewidth}
    \centering
    \includegraphics[width=\linewidth]{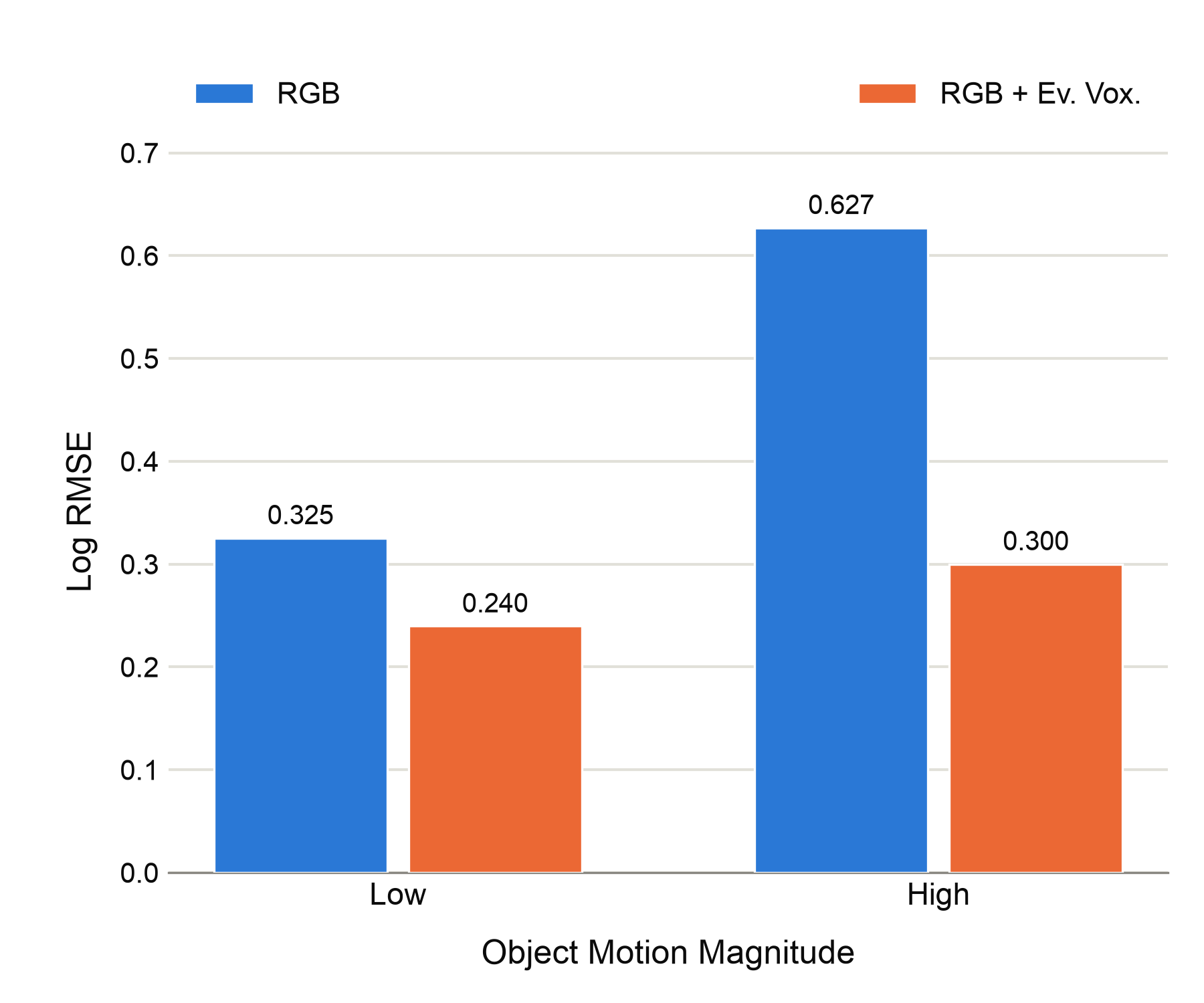}
    \caption{Object motion}
    \label{fig:sens-obj-motion}
\end{subfigure}
\caption{Sensitivity of reconstruction error to camera and object motion magnitude on a subset of the PointOdyssey test set with MonST3R. RGB + Ev.\ Vox.\ consistently achieves lower Log RMSE than RGB alone, with the gap widening under high object motion (Figure~\ref{fig:sens-obj-motion}). Camera motion is divided into three levels (Low, Medium, High) based on the ground-truth camera poses, and object motion is divided into two levels (Low, High) based on the ground-truth depth variation.}
\label{fig:sensitivity}
\end{figure}

\subsubsection{Event+RGB against Camera Motion.}
In Figure~\ref{fig:trajectories}, we showcase the trajectories predicted by the RGB + Ev. Vox. variant with MonST3R as the base model. We observe that smooth, uniform camera motion (Figure~\ref{fig:traj-shaman3}) is easier to estimate, while more non-uniform motion (Figure~\ref{fig:traj-ambush6}) accumulates higher error over time, resulting in inaccurate trajectories.

Furthermore, in Figure~\ref{fig:sens-cam-motion} we measure the sensitivity of reconstruction error to camera motion on a subset of the PointOdyssey test set with MonST3R. We divide the PointOdyssey test set into three levels of camera motion using the ground-truth camera poses. RGB and event voxel fusion achieves lower Log RMSE across all three levels of camera motion, underscoring the importance of event and RGB fusion for accurate 4D reconstruction.

\subsubsection{Event+RGB against Object Motion.}

In Figure~\ref{fig:sens-obj-motion}, we further measure the sensitivity of reconstruction error to object motion. In this sensitivity study, we divide the PointOdyssey test set into two levels of object motion based on ground-truth depth variation. While RGB and event voxel fusion achieves lower Log RMSE across both levels of object motion, gains are more pronounced under high object motion, achieving a 0.327 lower Log RMSE in this category. This further reinforces the importance of the complementary event and RGB modalities for accurate 4D reconstruction, particularly under high object motion.

\begin{figure}[htb]
\centering
\begin{subfigure}[b]{0.48\linewidth}
    \centering
    \includegraphics[width=\linewidth]{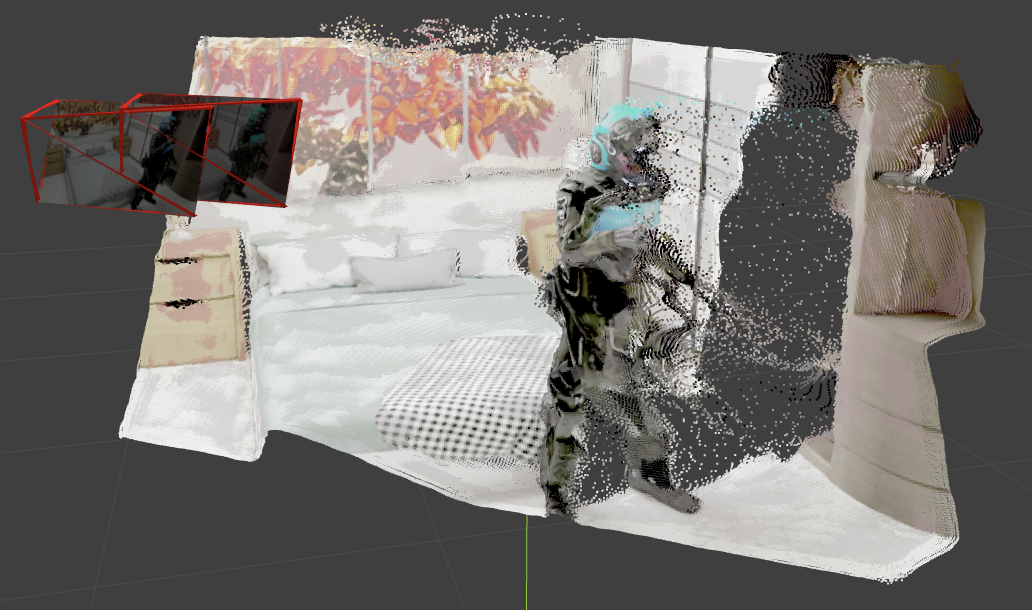}
    \caption{}
    \label{fig:qual-1}
\end{subfigure}
\hfill
\begin{subfigure}[b]{0.48\linewidth}
    \centering
    \includegraphics[width=\linewidth]{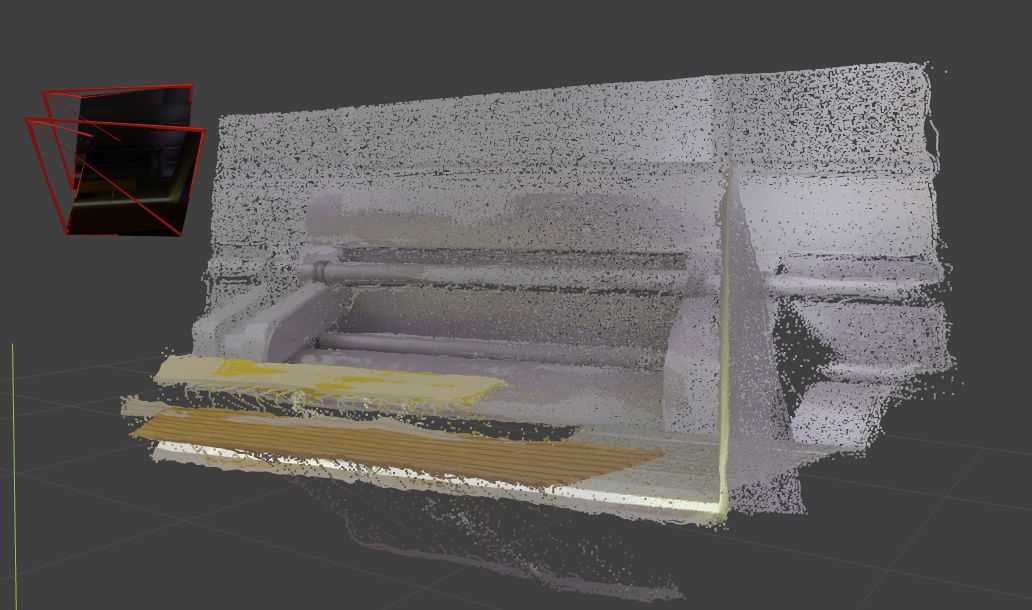}
    \caption{}
    \label{fig:qual-2}
\end{subfigure}
\\[0.5em]
\begin{subfigure}[b]{0.48\linewidth}
    \centering
    \includegraphics[width=\linewidth]{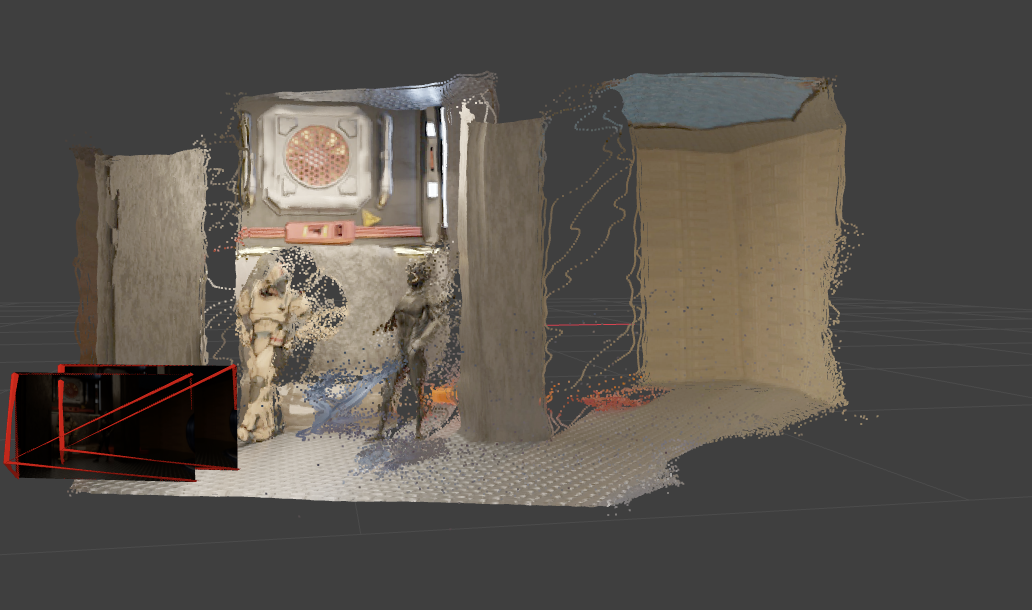}
    \caption{}
    \label{fig:qual-3}
\end{subfigure}
\hfill
\begin{subfigure}[b]{0.48\linewidth}
    \centering
    \includegraphics[width=\linewidth]{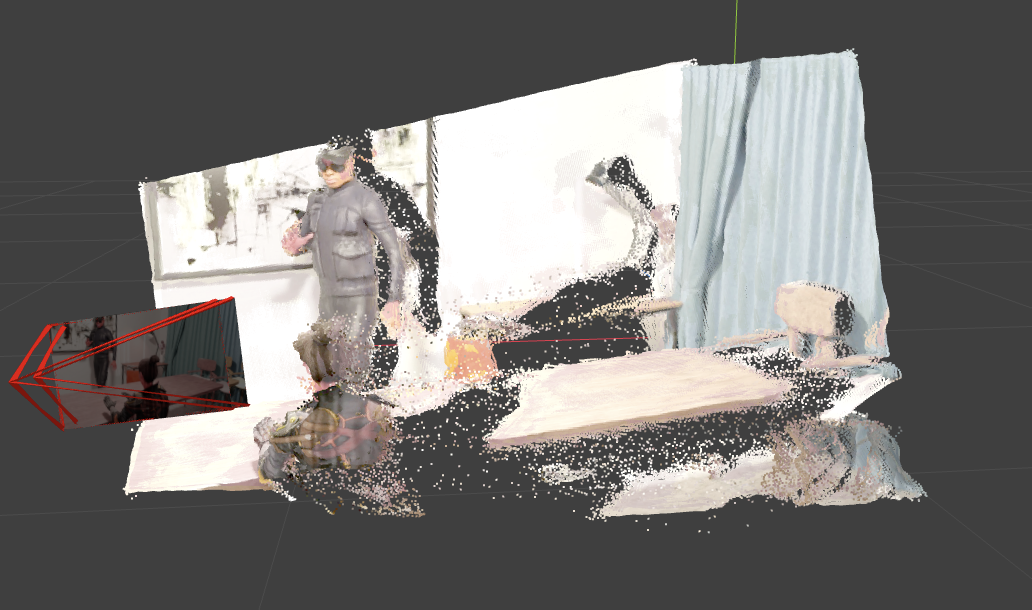}
    \caption{}
    \label{fig:qual-4}
\end{subfigure}
\caption{Qualitative 4D reconstructions produced by the RGB + Ev.\ Vox.\ variant with MonST3R as the base model.}
\label{fig:qual-pointclouds}
\end{figure}

\subsubsection{Visualization of RGB+Event Voxels based 4D Reconstruction.}

In Figure~\ref{fig:qual-pointclouds}, we showcase point cloud reconstructions produced by the RGB + Ev.\ Vox.\ variant, with MonST3R as the base model, from pairs of RGB frames and corresponding event voxels. Across these visualizations, we observe that scene geometry remains intact using our proposed fusion of RGB and event data for 4D reconstruction.

\section{Discussion \& Future Work}
\label{ref:discussion}

This preliminary study shows that a minimal event-camera branch added to MonST3R reduces both depth and pose estimation error across the benchmarks considered, with the largest gains under high camera and object motion. The design is deliberately simple: an event-voxel branch concatenated with RGB features, with no other architectural changes. This leaves several directions open for a more thorough follow-up study.

\begin{itemize}
    \item \textbf{Event-specific encoders.} We encode events as dense voxel grids processed by the same backbone used for RGB. Architectures designed for the sparse, asynchronous structure of event data, such as Spiking Neural Networks (SNNs) \cite{ghosh2009spiking} or event transformers \cite{sabater2022event}, may extract more informative motion cues than a standard RGB-based backbones.
    \item \textbf{Adaptive multi-modal fusion.} We fuse modalities by concatenation, which treats RGB and event features uniformly regardless of local motion. Events are inherently adaptive to motion, while RGB provides uniform spatial coverage independent of it; fusion mechanisms such as cross-attention or gating could weight each modality's contribution based on local motion cues rather than combining them uniformly.
    \item \textbf{More Recent 4D base models.} We build on MonST3R, which is no longer the most accurate feed-forward 4D reconstruction model. Applying the same event-branch extension to more recent architectures, such as VGGT \cite{wang2025vggt} or PAGE4D \cite{zhou2026page}, would test whether the benefit of event fusion generalizes across base models rather than being specific to MonST3R.
    \item \textbf{Real event data.} Sintel, Bonn, KITTI and PointOdyssey do not natively contain event streams, so the event modality used in our evaluation is synthetically generated using V2E simulator. Validating these gains on real event-camera captures is an important next step before drawing conclusions about real-world deployment.
\end{itemize}

\section{Conclusion}
\label{ref:conclusion}

We present a preliminary study on 4D reconstruction using a combination of RGB frames and event voxels. To this end, we propose a minimalist design that concatenates event voxels with RGB frames before passing them through MonST3R, retrained to incorporate this additional modality. Across multiple datasets for camera pose estimation and video depth estimation, we observe that combining event voxels and RGB frames achieves lower reconstruction error than either modality alone, indicating that spatial information from RGB and motion cues from event voxels play complementary roles in accurate 4D reconstruction. This preliminary work lays the foundation for more principled multi-modal fusion approaches combining events and RGB for 4D reconstruction, which we leave for future work.

\section{Acknowledgments}

This research was supported by the Singapore National Research Foundation (NRF), Prime Minister’s Office, Singapore, under its Campus for Research Excellence and Technological Enterprise (CREATE) programme. The Mens, Manus, and Machina (M3S) is an interdisciplinary research group (IRG) of the Singapore-MIT Alliance for Research and Technology (SMART) centre.

\bibliographystyle{splncs04}
\bibliography{main}
\end{document}